\documentclass[11pt]{article}
\usepackage{times} 
\usepackage[final]{graphicx}

\usepackage{float}
\usepackage[table]{xcolor}

\usepackage[most]{tcolorbox}
\newtcolorbox{highlighted}{colback=yellow,breakable}
\usepackage{lipsum}
\usepackage{url} 
\usepackage{graphicx} 
\usepackage[margin=1in]{geometry} 
\usepackage{authblk} 
\usepackage{amsmath} 
\usepackage{amsfonts}
\usepackage{tabularx,ragged2e,booktabs} 
\usepackage{titlesec}
\usepackage{amssymb}
\usepackage{comment}
\usepackage{subcaption}
\usepackage{authblk}
\usepackage{xcolor} 
\usepackage{hyperref}
\usepackage{amssymb}
\usepackage{dsfont}
\usepackage{float}
\usepackage[ruled,vlined]{algorithm2e}
\usepackage{multirow}
\usepackage[normalem]{ulem} 
\usepackage{graphicx,lipsum,wrapfig}
\usepackage{cleveref}
\usepackage[flushleft]{threeparttable}
\usepackage{booktabs,caption}
\usepackage[sorting=none,style=ieee,citestyle=numeric-comp, backend=biber]{biblatex}
\bibliography{bib}
\usepackage{nameref}

\usepackage{wrapfig}
\titleformat*{\subsubsection}{\normalfont}
\usepackage{color}

\begin{document}

\begin{titlepage}
    \begin{center}
        \vspace*{1cm}

        \LARGE{Neural Summarization of Electronic Health Records}
        \vspace*{1cm}
    \end{center}
    
    \noindent \large{\textbf{All Co-Authors:}}\vspace*{0.1cm}\\
    \noindent Koyena Pal$^1$\\
    \noindent Seyed Ali Bahrainian$^{1,3}$\\
    \noindent Laura Mercurio, MD$^{2,4}$\\
    \noindent Carsten Eickhoff$^{1,2,3}$\\

    \noindent \footnotesize{\textit{$^1$Department of Computer Science, Brown University, Providence, RI, USA}}\\
    \footnotesize{\textit{$^2$Faculty of Medicine, University of T\"{u}bingen, T\"{u}bingen, Germany}}\\
    \footnotesize{\textit{$^3$Institute for Bioinformatics and Medical Informatics, University of T\"{u}bingen, T\"{u}bingen, Germany}}\\
    \footnotesize{\textit{$^4$Departments of Pediatrics and Emergency Medicine, Alpert Medical School of Brown University, Providence, RI, USA}}\\

    \vspace*{1cm}
    
    \noindent \large{\textbf{Corresponding Authors:}}\vspace*{0.1cm}\\
    \noindent Carsten Eickhoff, PhD\\
    233 Richmond Street\\
    Providence, RI 02903\\
    carsten@brown.edu\\
    (401)-863-9665 \\
    \vspace*{0.25cm}
    
    \vspace*{1cm}
    \noindent \large{\textbf{Keywords:}} summarization; medical report summarization; electronic health record; dataset design; artificial intelligence; deep learning
    \vspace*{1cm}
    
    \vspace*{1cm}
    
    \noindent
    \noindent

\end{titlepage}

\author[1]{Koyena Pal}
\author[1]{Seyed Ali Bahrainian}
\author[1,2,*]{Carsten Eickhoff}

\affil[1]{Department of Computer Science, Brown University}
\affil[2]{Center for Biomedical Informatics, Brown University}
\affil[*]{Co-corresponding Authors}
\date{\vspace{-1.6cm}}

\title{Neural Summarization of Electronic Health Records}

\noindent

\section*{Abstract}

\subsection*{Background}
Electronic Health Record (EHR) summarization is the process of condensing and extracting relevant information from EHRs to provide key patient health details in a concise manner. Such summaries are helpful in improving healthcare efficiency and decision-making for healthcare providers.

\subsection*{Objective}
\textbf{Hospital discharge documentation is among the most essential, yet time-consuming documents written by medical practitioners}. The objective of this study was to automatically generate hospital discharge summaries using neural network summarization models. In particular, we studied various data preparation and neural network training techniques that generate discharge summaries.
\subsection*{Materials and Methods}
Using nursing notes and discharge summaries from the MIMIC-III dataset, 
we studied the viability of the automatic generation of various sections of a discharge summary using four state-of-the-art neural network summarization models (BART, T5, Longformer and FLAN-T5). 

\subsection*{Results}
Our experiments indicated that training environments including nursing notes as the source, and discrete sections of the discharge summary as the target output (e.g. ``History of Present Illness'') improve language model efficiency and text quality.
According to our findings, the fine-tuned BART model improved its ROUGE F1 score by 43.6\% against its standard off-the-shelf version. We also found that fine-tuning the baseline BART model with other setups caused different degrees of improvement (up to 80\% relative improvement). 
We also observed that a fine-tuned T5 generally achieves higher ROUGE F1 scores than other fine-tuned models and a fine-tuned FLAN-T5 achieves the highest ROUGE score overall, i.e., 45.6.

\subsection*{Discussion}
This study demonstrates the general viability of the automatic generation of parts of the discharge summary; a key step in reducing the clerical burden on healthcare providers. For majority of the fine-tuned language models, summarizing discharge summary report sections separately outperformed the summarization the entire report quantitatively. On the other hand, fine-tuning language models that were previously instruction fine-tuned showed better performance in summarizing entire reports.

\subsection*{Conclusion}
This study concludes that a focused dataset designed for the automatic generation of discharge summaries by a language model can produce coherent Discharge Summary sections.


\newpage

\section*{INTRODUCTION}

\begin{figure}[h!]
\begin{center}
  \includegraphics[width=0.9\columnwidth]{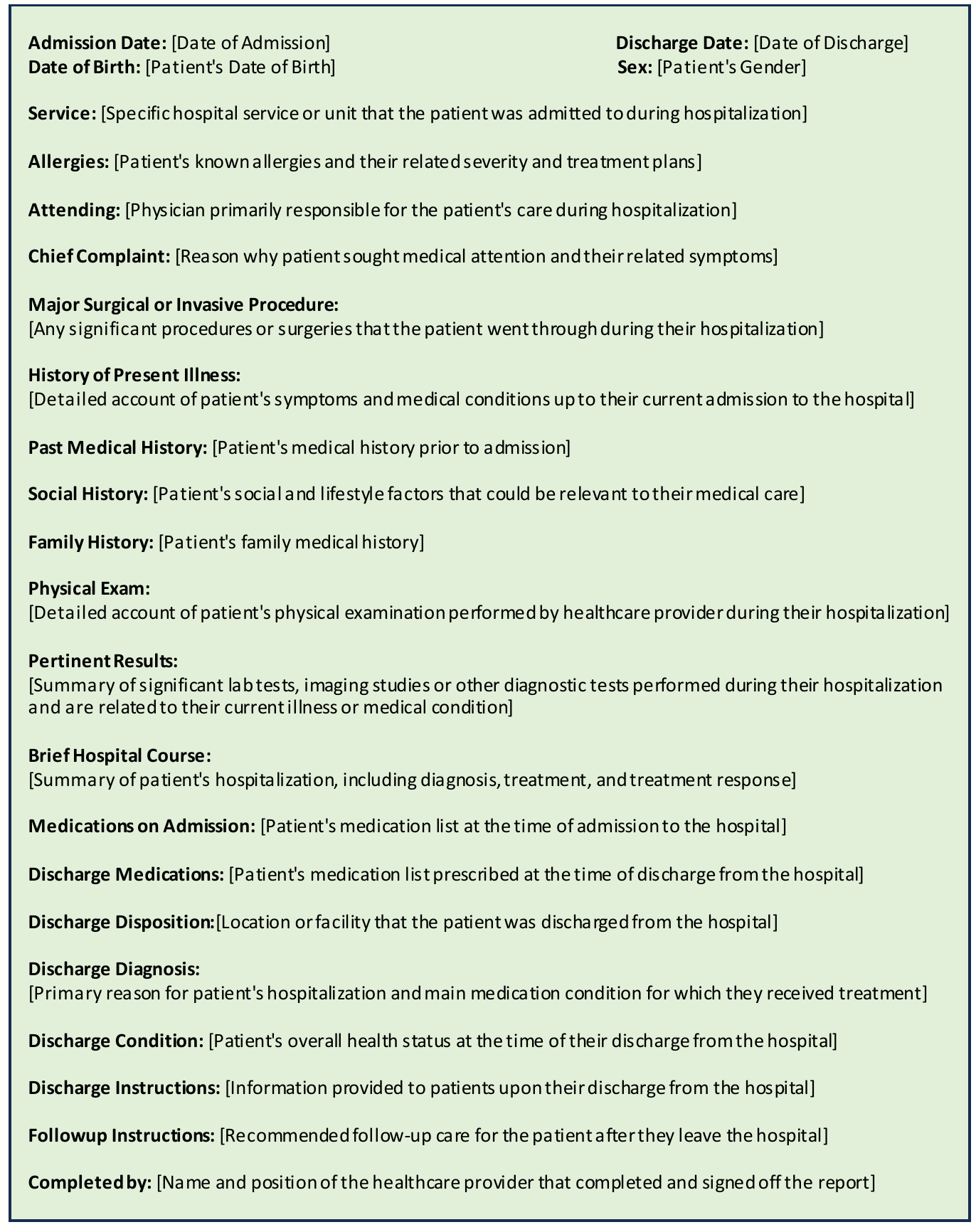}
  \caption{\textbf{Discharge Summary Report Template:} A general template of Discharge Summary Report present in the MIMIC-III dataset.}~\label{fig:dischargeSum}
 \end{center}
\end{figure}

When patients leave the hospital, their discharge summary represents a key document in the Electronic Healthcare Record (EHR) describing relevant medical conditions, events, and planned interventions/treatments during their stay, as shown in Figure ~\ref{fig:dischargeSum} and Table ~\ref{tab:eval1}. 
Together with various other EHR notes, discharge summaries provide key medical information to future providers. As such, it is unsurprising that multiple studies have shown that medical professionals spend at least twice as much time on EHR documentation than on patient care~\cite{Oxentenko2010, Holmgren2021ResidentPE}. At a system level, most hospitals face major challenges around digitally exchanging healthcare information between institutions as well as public health agencies ~\cite{healthit}. Common reported barriers include lack of capacity (e.g., technical support, staffing), interface-related issues (e.g. cost, complexity), vocabulary inconsistencies, and difficulty in extracting relevant information. To address these barriers, we investigate the effectiveness of modern natural language generation methods and data organization for fine-tuning and automatically generating various sections of a discharge summary report. This work represents a key step towards more efficient text summarization, and hopefully reducing the documentation burden faced by healthcare providers.


Previous studies indicate that pre-processing of raw, non-annotated datasets is more complex than that of annotated datasets ~\cite{GanesanAndSubotin2015, CarvalhoAndCurto2014, Diaz2020TowardsAG}. Annotated datasets generally include tags to identify concepts or human-written text summaries to describe the data content, while non-annotated datasets do not include human-generated labels. Since non-annotated datasets are more widely available, we utilize them by automatically creating annotated training and testing datasets for improving pre-trained language models. The language models used for text summarization usually have either of the two types of output – extractive summaries or abstractive summaries. The former is created using parts of sentences from original documents while the latter is based on the concepts captured, which is conveyed by the model using its vocabulary span. Neural Network language models are becoming increasingly popular for summarization tasks in both non-medical (e.g., general news ~\cite{newssum}) and  medical contexts (~\cite{Diaz2020TowardsAG, AlsentzerAndKim2018, CERC2020}). 

In this study, we compare various data creation setups for training language models and combine them with a variety of different deep learning text summarization techniques to identify the most robust settings in terms of ROUGE metrics for the automatic composition of discharge summary text. We focus on generating text using nursing notes as inputs because they have a notable overlap in content with the information present in the discharge summaries.

Concretely, we explore the following research questions:
\begin{enumerate}
    \item What models and their training setups are best for achieving high-quality medical text summaries?
    \item What parts of the discharge summary report can we automatically generate using nursing notes?
\end{enumerate}

\subsection*{Background and Significance}

\subsubsection{General Text Summarization}
In this section, we briefly review extractive and abstractive approaches that have been used in non-medical documents. It includes several models central to this paper. Recent works on text summarization algorithms can be broadly classified based on the type of summary generated – extractive and abstractive. Extractive summarization involves taking a subset of phrases and sentences from the input documents and concatenating them to form a summary. On the other hand, abstractive summarization algorithms produce summaries based on their own vocabulary and the concepts they associate with the input document, not necessarily using the exact words of a source article.

One of the models in the extractive summarization category is the Luhn summarizer~\cite{Luhn1958TheAC}. It selects sentences based on the maximum number of significant words present in a particular sentence. The significance of words is determined through Term Frequency - Inverse Document Frequency, also known as TF-IDF~\cite{AIZAWA200345}. A common baseline model for extractive summarization is LEAD-3, which is another extractive summarization solution that takes the first three sentences of the input document and sets them as the document's summary. Another sub-category of extractive summary algorithms is topic-based approaches. For example, the model designed by Harabagiu et al.~\cite{harabagiuetal} represents topic themes based on events that frequently occur over a set of documentation. They illustrate five ways of determining such frequencies – topic signatures, enhanced topic signatures, thematic signatures, modeling documents' content structure, and templates. There are also graph-based~\cite{Erkan2004LexRankGL, Baralis2013GraphSumDC} approaches that use text representation in a graph where words or sentences are represented as nodes and semantically-related text elements are connected through edges. Finally, discourse-based approaches~\cite{mannetal} integrate linguistic knowledge to represent the connections within and between sentences. 

A state-of-the-art model for abstractive text summarization is the Bidirectional and Auto-Regressive Transformer (BART)~\cite{bart}. As the name suggests, BART employs a standard Transformer-based neural machine translation architecture with a bidirectional encoder and a left-to-right decoder. Its encoder behaves similarly to Bidirectional Encoder Representations from Transformers (BERT), another well-known transformer~\cite{devlin-etal-2019-bert}, while its decoding nature resembles that of Generative Pre-trained Transformers (GPT)~\cite{Radford2018ImprovingLU, Radford2019LanguageMA, Radford2020}. Other similar recent language models include UniLM~\cite{unilm}, MASS~\cite{mass}, and XLNet~\cite{xlnet}. Apart from BART, other popular models include Text-To-Text Transfer Transformer (T5)~\cite{t5} and Longformer~\cite{longformer}. T5 is an encoder-decoder model pre-trained on various text-based language tasks, with input-output definitions converted into a text-to-text format. Longformer also has a transformer architecture; however, it is modified to process lengthy document texts. FLAN-T5~\cite{chung2022scaling} is an instruction-fine tuned T5 model. It is recent state-of-the-art model that was trained on 1000 additional tasks compared to the T5 model. Using these models, we aim to efficiently generate discrete segments of medical discharge summaries from nursing notes using abstractive summarization. 

In this study, we compare four representative models: BART, T5, FLAN-T5, and Longformer. We employed the BART framework since it has shown superior summarization performance on a number of non-medical benchmark datasets. The T5 model was selected as a candidate because it is an early seq2seq Transformer model which has been trained on a large amount of data for multiple natural language processing applications including summarization. We test the FLAN-T5 model since it has been trained on even more number of tasks than T5. We include both models because we are curious to understand how exactly a large language model such as FLAN-T5 is better than a language model like T5 when applied to the same tasks. Finally, we tested the Longformer because it distinguishes itself from the most Transformer-based models in that its capable of handling also lengthy texts.

\subsubsection{EHR Summarization}

Due to information overload and time involved in preparing and utilizing EHR documents~\cite{infooverload1, contentclinicalprob, challengesmedicalrecords}, healthcare providers report shrinking amounts of interaction with their patients. As a result, there has been an ongoing push for automated integration of clinical reports to produce detailed, yet concise, medical summaries~\cite{adams2021whats, adams2022learning}. Current approaches within EHR summarization have mostly been extractive in nature, in which summarized text is directly taken from the original medical document. This approach ranges from extracting relevant sentences from the input text to create a summary ~\cite{Moen2016ComparisonOA}, topic modeling using Bayesian network or Latent Dirichlet allocation ~\cite{Pivovarov2015ElectronicHR}, creating heuristic rules ~\cite{GOLDSTEIN2016159}, to utilizing neural networks ~\cite{AlsentzerAndKim2018, Kanwal_2022}. 

On the other hand, there have been works on medical document summarization that are abstractive in nature~\cite{10.1016/j.csl.2021.101276}. Zhang et al.~\cite{initradio}, for instance, utilized the findings and background section from chest x-ray radiology reports to generate an assessment section. MacAvaney et al.~\cite{MacAvaneyetal} furthered this framework by including the encoding of an ontology report section to aid the decoding process. By doing so, they were able to create an ontology-aware clinical abstractive summarization model. 

To capture the complexity of long medical texts, while retaining the ability to generate abstract summaries, recent studies combined these techniques to create an extractive-abstractive pipeline for summarization. Shing et al.~\cite{shing2021}, for instance, uses a recall-oriented extractor to extract relevant sentences and then an abstractor component to remove irrelevant or duplicated information. Our dataset design setups include a similar pipeline. Instead of having a two component pipeline while summarizing an input text, we utilize raw clinical documents and an extractive approach to create source-target pairs. By doing so, we aim to train the main one-component abstractive model (BART, T5, FLAN-T5, or Longformer) to identify key sections of the source text to produce the intended target pair with additional information the model deems relevant from the nursing notes. There is a very recent work by Searle et. al~\cite{Searle_2023} that focuses on generating Brief Hospital Course (BHC) text, which is a sub-section in the Discharge Summary Report. They use a novel ensemble model that incorporates a medical concept ontology and show that this model outperforms their baseline models. Our work looks into both partial and full discharge summarization with extractive-abstractive summarization pipelines (Setups 2 and 3 described in the following section) as well as only abstractive schemes.

\section*{METHODS AND MATERIALS}
\subsection*{Data Description}

\begin{figure}[h!]
\begin{center}
  \includegraphics[width=\columnwidth]{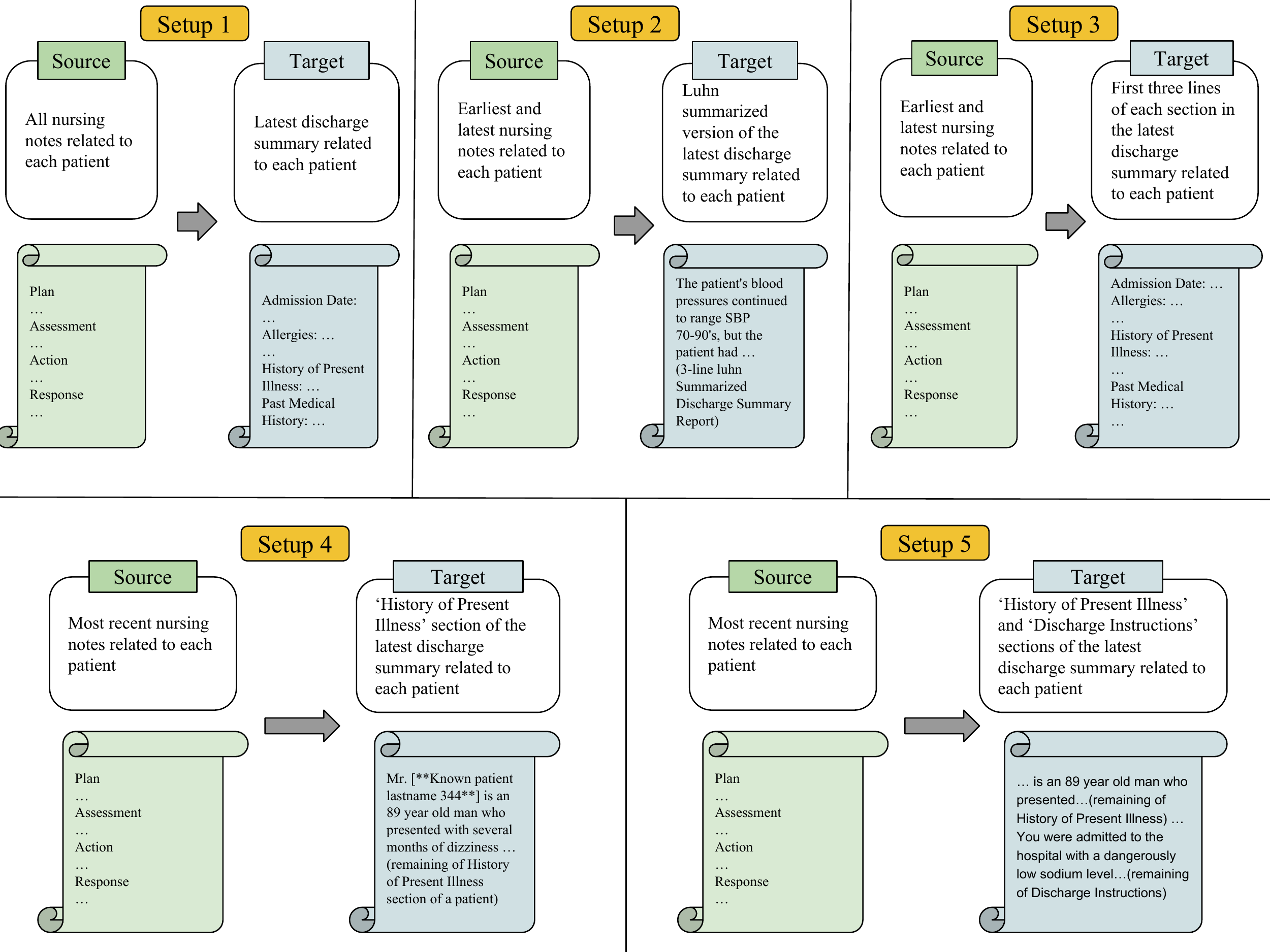}
  \caption{\textbf{Training Dataset Creation Overview}: Within each setup, there is a source-target pair for each patient in the training data. The figures describe how each source-target pair is created with an example of how they look like.}~\label{fig:datasetOverview}
 \end{center}
\end{figure}

The data for this study is from the Medical Information Mart for Intensive Care (MIMIC-III) database~\cite{johnson2016mimic}. This relational database contains de-identified health data for over 40,000 patients admitted to critical care units in the Beth Israel Deaconess Medical Center between 2001 and 2012. We used the ``noteevents'' table, which contains notes written by a wide-range of healthcare providers. There are 15 types of notes present, including hospital discharge summaries, echocardiography reports, electrocardiography (ECG) interpretations, nursing notes, and many more. Among them, we focused on discharge summaries and nursing notes. We selected nursing notes as model inputs because they empirically demonstrated a substantial overlap in content with the information present in the discharge summaries. We then designed the following setups for training text summarization models:
\begin{itemize}
    \item \underline{Setup 1:} For each patient, we gathered all the nursing notes and placed them together under the ‘source’ column. The corresponding 'target' was set to be the most recent discharge summary for that patient.
    \item \underline{Setup 2:} For each patient, we combined the earliest and latest nursing notes to represent the ‘source’ part of the source-target pair. To create the ‘target,’ we used the Luhn summarizer (reviewed in Section 2) on the patient's most recent discharge summary and set the generated texts as the respective ‘target.’
    \item \underline{Setup 3:} The source setup is the same as setup 2. For the ‘target,’ we extracted the first three lines of each section within the most recent discharge summary report.
    \item \underline{Setup 4:} For each patient, the single most recent nursing note was considered as ‘source,’ and the ‘History of Present Illness’ section in their most recent discharge summary was considered as ‘target.’
    \item \underline{Setup 5:} This setup is similar to setup 4, except that we included the ‘History of Present Illness’ as well as the ‘Discharge Instructions’ sections as part of the respective ‘target’ text.
\end{itemize}

Setups 1, 2, and 3 include $6,157$ training data points, while setups 4 and 5 contain $6,132$ and $5,981$ training data points, respectively. For testing purposes, we withheld $1,000$ patient data reports for each setup. The Python code written to generate these setups is available here\footnote{Setup Links: \href{https://colab.research.google.com/drive/1GDSfGyw_MlzYYbmapFUItXdh2KcNn64c?usp=sharing}{part 1}, \href{https://colab.research.google.com/drive/130mfrPZ_mamI8ijyacOSwo225Jx5dNva?usp=sharing}{part 2}}. The purpose of creating these five setups is to understand the input combination best understood by the models with respect to the expected output; a partial discharge summary.

Setup 1 aims at using all the existing data, i.e., the entire history of nursing notes to generate a discharge summary report. This setup involves pre-processing in order to create the training dataset, meaning that the source includes all the raw nursing notes for a set of patients, and the target includes the discharge summary reports for the same set of patients. It features examining the usefulness of the entire history of nursing notes for each patient. At the same time, the generation target is the entire discharge summary.

Setups 2 and 3 aim at augmenting the dataset with target summaries as opposed to the entire discharge summary, as an attempt to generate a short summary.
In these setups, additional processing is required to populate both the source and the target text. These setups use the first and last nursing notes for each patient as the source document. In setup 2, the target summary is generated using the Luhn  summarizer~\cite{Luhn1958TheAC}, while Setup 3 uses a modified version of the LEAD-3 pipeline, which pulls the first three lines of each section in the discharge summary report rather than the first three lines of the entire document. Both of these summarizers represent extractive summarization algorithms. Therefore, Setups 2 and 3 evaluate whether a combination of extractive and abstractive pipelines can generate better summary reports than a purely abstractive approach.

Setups 4 and 5 aim at narrowing their source texts to only the most recent nursing note. This is based on the assumption that the last nursing note is the most recent note which contains the latest and most relevant information regarding discharging a patient. Then, for Setup 4 the target output is one section of the discharge summary report, namely, `History of Present Illness.' On the other hand,  Setup 5 focuses on automatically generating two sections of the discharge summary report as the target, namely, `History of Present Illness', as well as, the `Discharge Instructions'. The two sections of a discharge summary are the most text-rich sections, therefore, generating them may be easier automated using a language model.

\subsection*{Metrics}
To quantitatively assess the summaries generated by each setup and model combination, we use the popular Recall-Oriented Understudy for Gisting Evaluation (ROUGE) metrics, ROUGE-1, ROUGE-2, ROUGE-L, and ROUGE-L-SUM~\cite{lin-2004-rouge}. ROUGE-1 measures the number of matching unigrams between the generated text and the actual text. Meanwhile, ROUGE-2 measures the number of matching bigrams instead of unigrams.  Instead of finding fixed n-gram matches, ROUGE-L computes the longest common subsequence (LCS) - i.e., the number of the longest shared word sequence, between the model output and the actual text. ROUGE-L-SUM makes this calculation and then applies it on a summary level. This means it splits the sentences in the text based on newline characters and takes the union LCS matches between actual (reference) text and every model-generated sentence. For these ROUGE scores, we calculate the precision and recall of the overlapped words between the generated and the original discharge summaries. We then measure $F_1$ scores calculated using the following formula:
\begin{equation}
    F1\_score = \frac{2*Precision*Recall}{Precision + Recall}
\label{eq:f1}
\end{equation}

In Tables ~\ref{tab:setup123results} and ~\ref{tab:setup4and5results}, we include two bold rows that are populated with ROUGE values. We display ROUGE scores, namely the precision, recall, and $F_1$ scores for all setup and model combinations. Each value is approximated to two decimal places. The first bold row showcases the highest scores achieved in the given setup. The second bold row represents the average values of ROUGE scores for each metric. This is calculated by taking the average over all the mid scores recorded in all the models, i.e., the pre-trained and fine-tuned versions of the BART, Longformer, T5, and FLAN-T5 models. In order to verify our findings' robustness, we conduct statistical significance tests based on bootstrap re-sampling using the official ROUGE package~\cite{lin-2004-rouge}.

\subsection*{Models}

To automate the text generation of discharge summary reports, we selected 4 text summarization models that would be trained and tested under the aforementioned setups. The models that are used for this purpose include BART~\cite{bart}, T5~\cite{t5}, Longformer~\cite{longformer} and FLAN-T5~\cite{chung2022scaling}. The rationale for the above model selection is: 
(1) BART represents the state of the art in abstractive summarization on different benchmark datasets among a few domains. Since nursing notes often have overlapping but not complete information, the selected summarization models must extract certain key information and simultaneously understand the overall idea described in the nursing notes. Analogous to the other two models, it is built by using a decoder similar to GPT~\cite{Radford2019LanguageMA} and an encoder similar to BERT~\cite{devlin-etal-2019-bert}. However, it differentiates itself from the other two models by adding noise to the input data which has shown to help the model perform better in summarization tasks.
(2) T5 represents a unified transferred learning framework that has been trained on very large datasets. Consequently, T5 can more easily adapt to most tasks than other models, such as text generation.  T5 is therefore a preferred model to train and test on a language-based task such as EHR summarization.
(3) The Longformer model can encode and decode multiple lengthy text documents. The  previously-mentioned transformer-based models have a length limitation due to the quadratic increase in scale and complexity caused by their attention mechanisms. However, the Longformer model is specifically designed to process long documents.
(4) FLAN-T5 represents an enhanced version of T5 since it is fine-tuned on more than 1000 additional tasks covered in more languages compared to T5. It is a recent open-source large language model that is comparable to performance of GPT ~\cite{Radford2019LanguageMA} models.
\section*{RESULTS}

\subsection*{Quantitative Results}

\begin{table}[]
\resizebox{\textwidth}{!}{%
\begin{tabular}{|c|l|l|l|l|l|}
\hline
\textbf{Setup No.} &
  \multicolumn{1}{c|}{\textbf{Model}} &
  \multicolumn{1}{c|}{\textit{\textbf{rouge1}}} &
  \multicolumn{1}{c|}{\textit{\textbf{rouge2}}} &
  \multicolumn{1}{c|}{\textit{\textbf{rougeL}}} &
  \multicolumn{1}{c|}{\textit{\textbf{rougeLsum}}} \\ \hline
\multirow{9}{*}{\begin{tabular}[c]{@{}c@{}}(1)\\ Source: \\ All NN\\ Training target: \\ Most recent DS\\ Testing target: \\ Most recent DS\end{tabular}} &
  BART (pre-trained) &
  32.86 / 9.90 / 14.43 &
  17.22 / 5.16 / 7.60 &
  28.45 / 8.42 / 12.35 &
  28.39 / 8.40 / 12.33 \\ \cline{2-6} 
 & BART (fine-tuned)             & 49.34 / 32.96 / 36.11          & 31.09 / 21.67 / 23.42          & 45.08 / 30.08 / 32.94          & 45.09 / 30.07 / 32.89          \\ \cline{2-6} 
 & Longformer (pre-trained)      & 11.97 / 4.23 / 5.60            & 5.72 / 1.76 / 2.40             & 10.62 / 3.68 / 4.89            & 10.62 / 3.67 / 4.87            \\ \cline{2-6} 
 & Longformer (fine-tuned)       & 37.73 / 25.10 / 27.30          & 20.30 / 13.93 / 14.95          & 31.84 / 20.06 / 21.92          & 31.80 / 20.89 / 22.72          \\ \cline{2-6} 
 & T5 (pre-trained)              & 24.13 / 5.21 / 8.23            & 10.02 / 2.12 / 3.41            & 20.70 / 4.40 / 6.99            & 20.64 / 4.39 / 6.97            \\ \cline{2-6} 
 & T5 (fine-tuned)               & 53.81 / 32.30 / 38.31          & 33.67 / 19.97 / 23.84          & 49.08 / 29.38 / 34.90          & 49.06 / 29.33 / 34.88          \\ \cline{2-6} 
 & FLAN-T5 (pre-trained)         & 25.03 / 7.69 / 10.93           & 11.56 / 3.67 / 5.31            & 21.74 / 6.48 / 9.33            & 21.65 / 6.47 / 9.31            \\ \cline{2-6} 
 & \textbf{FLAN-T5 (fine-tuned)} & \textbf{55.13 / 42.85 / 45.55} & \textbf{39.09 / 30.28 / 32.26} & \textbf{51.25 / 39.80 / 42.36} & \textbf{51.23 / 39.82 / 42.35} \\ \cline{2-6} 
 & \textbf{Average Scores}       & \textbf{36.25 / 20.03 / 23.31} & \textbf{21.08 / 12.32 / 14.15} & \textbf{32.35 / 17.79 / 20.71} & \textbf{32.31 / 17.88 / 20.79} \\ \hline
\multirow{9}{*}{\begin{tabular}[c]{@{}c@{}}(2)\\ Source: \\ Earliest + Latest NN\\ Training target: \\ Luhn Summarized DS\\ Testing target:\\ Most recent DS\end{tabular}} &
  BART (pre-trained) &
  31.49 / 9.99 / 14.19 &
  16.79 / 5.28 / 7.67 &
  27.17 / 8.46 / 12.16 &
  27.14 / 8.46 / 12.16 \\ \cline{2-6} 
 & BART (fine-tuned)             & 33.72 / 16.33 / 20.98          & 19.25 / 9.04 / 11.86           & 27.98 / 13.25 / 18.26          & 27.95 / 13.23 / 17.25          \\ \cline{2-6} 
 & Longformer (pre-trained)      & 11.58 / 4.13 / 5.49            & 5.32 / 1.73 / 2.34             & 10.15 / 3.57 / 4.75            & 10.16 / 3.58 / 4.75            \\ \cline{2-6} 
 & Longformer (fine-tuned)       & 26.90 / 15.36 / 18.20          & 12.33 / 6.51 / 8.02            & 21.07 / 11.26 / 14.33          & 21.07 / 11.27 / 13.80          \\ \cline{2-6} 
 & T5 (pre-trained)              & 22.97 / 5.00 / 7.81            & 9.51 / 1.95 / 3.13             & 19.55 / 4.17 / 6.56            & 19.60 / 4.18 / 6.56            \\ \cline{2-6} 
 & \textbf{T5 (fine-tuned)}      & \textbf{38.75 / 16.53 / 22.30} & \textbf{22.00 / 9.14 / 12.51}  & \textbf{32.06 / 13.53 / 18.35} & \textbf{32.13 / 13.56 / 18.39} \\ \cline{2-6} 
 & FLAN-T5 (pre-trained)         & 28.90 / 9.00 / 12.87           & 14.46 / 4.57 / 6.66            & 24.81 / 7.56 / 10.90           & 24.79 / 7.57 / 10.90           \\ \cline{2-6} 
 & FLAN-T5 (fine-tuned)          & 30.91 / 12.56 / 16.72          & 16.38 / 6.58 / 9.00            & 26.30 / 10.53 / 14.14          & 26.37 / 10.55 / 14.18          \\ \cline{2-6} 
 & \textbf{Average Scores}       & \textbf{28.15 / 11.11 / 14.82} & \textbf{14.51 / 5.60 / 7.65}   & \textbf{23.64 / 9.04 / 12.43}  & \textbf{12.25 / 9.05 / 12.25}  \\ \hline
\multirow{9}{*}{\begin{tabular}[c]{@{}c@{}}(3)\\ Source: \\ Earliest + Latest NN\\ Training target: \\ LEAD-3 DS\\ Testing target: \\ Most recent DS\end{tabular}} &
  BART (pre-trained) &
  31.37 / 9.97 / 14.19 &
  16.77 / 5.27 / 7.63 &
  27.19 / 8.45 / 12.15 &
  27.10 / 8.45 / 12.14 \\ \cline{2-6} 
 & BART (fine-tuned)             & 66.69 / 13.47 / 22.10          & 57.19 / 10.91 / 18.07          & 63.89 / 12.99 / 21.28          & 63.85 / 12.99 / 21.29          \\ \cline{2-6} 
 & Longformer (pre-trained)      & 11.59 / 4.15 / 5.50            & 5.31 / 1.73 / 2.35             & 10.13 / 3.58 / 4.76            & 10.15 / 3.58 / 4.75            \\ \cline{2-6} 
 & Longformer (fine-tuned)       & 66.44 / 13.46 / 22.07          & 56.95 / 10.87 / 18.00          & 63.53 / 12.93 / 21.18          & 63.60 / 12.95 / 21.22          \\ \cline{2-6} 
 & T5 (pre-trained)              & 22.97 / 5.00 / 7.81            & 9.51 / 1.95 / 3.13             & 19.55 / 4.17 / 6.56            & 19.60 / 4.18 / 6.56            \\ \cline{2-6} 
 & T5 (fine-tuned)               & 63.40 / 13.10 / 21.28          & 53.14 / 10.27 / 16.90          & 60.42 / 12.51 / 20.31          & 60.47 / 12.54 / 20.36          \\ \cline{2-6} 
 & FLAN-T5 (pre-trained)         & 28.89 / 9.02 / 12.88           & 14.52 / 4.58 / 6.68            & 24.83 / 7.57 / 10.91           & 24.82 / 7.58 / 10.90           \\ \cline{2-6} 
 & FLAN-T5 (fine-tuned)          & \textbf{71.35} / 10.25 / 15.73          & \textbf{61.86} / 6.83 / 10.53           & \textbf{68.34} / 9.32 / 14.40           & \textbf{68.36} / 9.29 / 14.35           \\ \cline{2-6} 
 & \textbf{Average Scores}       & \textbf{45.34 / 9.80 / 15.20}  & \textbf{34.41 / 6.55 / 10.41}  & \textbf{42.24 / 8.94 / 13.94}  & \textbf{42.24 / 8.95 / 13.95}  \\ \hline
\end{tabular}%
}
\caption{Mid range scores of ROUGE achieved by each model in data setups 1,2, and 3. NN stands for Nursing Notes and DS stands for Discharge Summary. The values in each cell represent the following measure: \{precision / recall / F1-measure\}.}
\label{tab:setup123results}
\end{table}

In Tables ~\ref{tab:setup123results} and ~\ref{tab:setup4and5results}, we evaluate the summaries generated by the four models, i.e. BART, T5, Longformer, and FLAN-T5, in their pre-trained and fine-tuned versions. The term ``pre-trained version'' signifies loading a pre-trained model (for instance, t5-base for T5), while the term ``fine-tuned version'' implies that we load a pre-trained model, train it further using our training dataset for a given data setup, and then finally test the newly trained model. 

For Setup 1, Table~\ref{tab:setup123results} indicates that a FLAN-T5 fine-tuned model achieves the highest scores in the ROUGE metric set. To understand whether this result is significant, we compare the FLAN-T5 fine-tuned version’s low-percentile $F_1$ score against the high-percentile $F_1$ scores achieved by other models. We find that all these values are significantly higher than all the other models. Between T5 fine-tuned and BART fine-tuned models, the difference in the former’s low-percentile scores and the latter’s high-percentile scores is not significant at ~0.1. This behavior is also reflected in Setups 2. Table~\ref{tab:setup123results} reflects the scores achieved by the models when they are trained and tested using Setup 2 data setup. Instead of FLAN-T5's fine-tuned version, T5’s fine-tuned version achieves the highest scores in Setups 2, which are significant compared to all other models except for BART’s fine-tuned version. This behavior is reflected in Setup 5 as well as shown in table ~\ref{tab:setup4and5results}. It means that T5 or BART can be applied to achieve comparatively high and similar ROUGE scores for these data setups.

For Setup 3, Table~\ref{tab:setup123results} indicates that BART fine-tuned model achieves the highest scores in the ROUGE metric set. This result is significant against all pre-trained versions and FLAN-T5 fine-tuned version. T5 and Longformer fine-tuned models achieve similar results, and the high-percentile $F_1$ ROUGE scores of these models are higher than the low-percentile $F_1$ ROUGE scores of BART’s fine-tuned version. This shows that we can interchangeably use any of T5, BART, and Longformer fine-tuned models to achieve similar ROUGE scores in this setup.

For Setup 4, Table~\ref{tab:setup4and5results} indicates that BART fine-tuned model also achieves the highest scores in the ROUGE metric set. Similar to Setup 3, this result is significant against all pre-trained versions and mildly significant against the T5 fine-tuned version. However, unlike Setup 3, this model ROUGE values are not significant against FLAN-T5's but are significant against Longformer for ROUGE-1, ROUGE-L, and ROUGE-L-SUM scores, but not in terms of ROUGE-2 scores. Since the majority of scores are significant, the BART fine-tuned model scores can be considered significantly higher than Longformer’s fine-tuned version. Hence, we can utilize T5, FLAN-T5, or BART to gain high ROUGE scores whilst generating a section of the discharge summary report using Setup 4.

In addition to testing model performance for each setup, we evaluate the ease with which models train and generate summary text by comparing the average $F_1$ values attained in each setup. Amongst Setups 1, 2, and 3 (full discharge summary output), Setup 2 achieves the lowest while Setup 1 achieves the highest. Between Setups 4 and 5 (partial discharge summary output) the difference between each setup ROUGE $F_1$ scores are not much different. Hence, both have similar ease of training and summarizing medical texts.

\begin{table}[]
\resizebox{\columnwidth}{!}{%
\begin{tabular}{|c|l|l|l|l|l|}
\hline
\textbf{Setup No.} &
  \multicolumn{1}{c|}{\textbf{Model}} &
  \multicolumn{1}{c|}{\textit{\textbf{rouge1}}} &
  \multicolumn{1}{c|}{\textit{\textbf{rouge2}}} &
  \multicolumn{1}{c|}{\textit{\textbf{rougeL}}} &
  \multicolumn{1}{c|}{\textit{\textbf{rougeLsum}}} \\ \hline
\multirow{9}{*}{\begin{tabular}[c]{@{}c@{}}(4)\\ Source: \\ Latest NN\\ Training target: \\ "History of Present Illness" \\ section of most recent DS\\ Testing target: \\ "History of Present Illness" \\ section of most recent DS\end{tabular}} &
  BART (pre-trained) &
  30.58 / 26.24 / 26.38 &
  18.70 / 16.39 / 16.59 &
  28.58 / 24.47 / 24.64 &
  28.57 / 24.44 / 24.61 \\ \cline{2-6} 
 & \textbf{BART (fine-tuned)} & \textbf{37.49 / 38.11 / 35.98} & \textbf{23.34 / 23.45 / 22.44} & \textbf{34.89 / 35.54 / 33.56} & \textbf{34.89 / 35.54 / 33.56} \\ \cline{2-6} 
 & Longformer (pre-trained)   & 12.83 / 12.89 / 10.75          & 7.17 / 7.08 / 5.90             & 12.32 / 12.33 / 10.31          & 12.29 / 12.37 / 10.30          \\ \cline{2-6} 
 & Longformer (fine-tuned)    & 26.27 / 31.61 / 27.11          & 11.16 / 13.17 / 11.35          & 23.53 / 28.48 / 24.35          & 23.53 / 28.47 / 24.36          \\ \cline{2-6} 
 & T5 (pre-trained)           & 20.76 / 12.10 / 14.12          & 10.05 / 5.75 / 6.88            & 18.73 / 10.97 / 12.78          & 18.80 / 10.97 / 12.79          \\ \cline{2-6} 
 & T5 (fine-tuned)            & 37.53 / 36.55 / 35.23          & 23.07 / 22.43 / 21.83          & 34.91 / 34.07 / 32.82          & 34.90 / 34.07 / 32.83          \\ \cline{2-6} 
 & FLAN-T5 (pre-trained)      & 26.41 / 22.67 / 22.51          & 15.37 / 13.71 / 13.74          & 24.28 / 20.81 / 20.74          & 24.34 / 20.92 / 20.80          \\ \cline{2-6} 
 & FLAN-T5 (fine-tuned)       & 37.20 / 35.15 / 34.41          & 22.59 / 21.86 / 21.37          & 34.42 / 32.71 / 31.94          & 34.35 / 32.67 / 31.89          \\ \cline{2-6} 
 & \textbf{Average Scores}    & \textbf{28.64 / 26.92 / 25.81} & \textbf{16.43 / 15.48 / 15.01} & \textbf{26.46 / 24.92 / 23.89} & \textbf{26.46 / 24.93 / 23.89} \\ \hline
\multirow{9}{*}{\begin{tabular}[c]{@{}c@{}}(5)\\ Source: \\ Latest NN\\ Training target: \\ "History of Present Illness" + "Discharge Instructions" \\ sections of most recent DS\\ Testing target: \\ "History of Present Illness" + "Discharge Instructions"\\  sections of most recent DS\end{tabular}} &
  BART (pre-trained) &
  31.28 / 26.29 / 26.61 &
  18.95 / 16.46 / 16.72 &
  29.38 / 24.72 / 25.07 &
  29.31 / 24.60 / 25.01 \\ \cline{2-6} 
 & BART (fine-tuned)          & 39.30 / 40.01 / 37.86          & 24.66 / 24.33 / 23.64          & 36.64 / 37.23 / 35.34          & 36.70 / 37.33 / 35.40          \\ \cline{2-6} 
 & Longformer (pre-trained)   & 12.62 / 12.92 / 10.61          & 7.02 / 7.18 / 5.81             & 12.05 / 12.38 / 10.12          & 12.04 / 12.36 / 10.12          \\ \cline{2-6} 
 & Longformer (fine-tuned)    & 25.29 / 25.29 / 22.20          & 12.51 / 11.51 / 10.53          & 23.21 / 22.88 / 20.23          & 23.19 / 22.90 / 20.21          \\ \cline{2-6} 
 & T5 (pre-trained)           & 22.04 / 13.20 / 15.41          & 10.73 / 6.80 / 7.87            & 20.10 / 12.19 / 14.18          & 20.10 / 12.18 / 14.18          \\ \cline{2-6} 
 & \textbf{T5 (fine-tuned)}   & \textbf{40.77 / 40.67 / 38.80} & \textbf{26.67 / 26.05 / 25.38} & \textbf{38.25 / 38.06 / 36.42} & \textbf{38.27 / 38.17 / 36.51} \\ \cline{2-6} 
 & FLAN-T5 (pre-trained)      & 22.81 / 19.81 / 19.41          & 12.48 / 11.43 / 11.23          & 20.83 / 18.04 / 17.78          & 20.87 / 18.07 / 17.80          \\ \cline{2-6} 
 & FLAN-T5 (fine-tuned)       & 38.46 / 35.35 / 34.91          & 24.51 / 22.66 / 22.64          & 35.85 / 33.00 / 32.59          & 35.87 / 32.97 / 32.62          \\ \cline{2-6} 
 & \textbf{Average Scores}    & \textbf{29.07 / 26.69 / 25.73} & \textbf{17.19 / 15.80 / 15.48} & \textbf{27.04 / 24.81 / 23.97} & \textbf{27.04 / 24.82 / 23.98} \\ \hline
\end{tabular}%
}
\caption{Mid range ROUGE scores achieved by each model in data setups 4 and 5. NN stands for Nursing Notes and DS stands for Discharge Summary. The values in each cell represent the following measure: \{precision / recall / F1-measure\}.}
\label{tab:setup4and5results}
\end{table}

\begin{table*}
  \centering
  \begin{tabular}{p{0.5\textwidth} | p{0.4\textwidth}} 
    \toprule
    Actual Discharge Summary Report & Generated Text \\
    \midrule
    \shortstack[l]{
    \textbf{Service:} MEDICINE \\
\textbf{Allergies:} Penicillins / Latex / Sulfa \\
(Sulfonamide Antibiotics) / 
Shellfish Derived \\
\textbf{Major Surgical or Invasive Procedure:}\\
s/p ERCP with metal stent placement [**2907-3-3**].\\
\\
\textbf{History of Present Illness:} \\
Ms. [**Known patient lastname 35261**] is a 77 \\
year old female with pancreatic cancer with liver \\ 
and lung metastases with recent failure to \\
gemcitabine treatment who presents with new \\
onset jaundice two days ago. She reports feeling \\
fatigued and having a poor a[** Location **] ie \\
ovr the last three days, and noted jaundice on \\
the morning of [**2907-2-28**]. She denies fevers, \\
but does report night sweats that have been \\
going on for weeks. She had one episode of \\
nausea and vomiting after taking a pain pill on \\
the evening of the [**2907-3-1**]. She \\ 
reports [**Location (un) 685**] colored \\
stool but denies pruritis or dark colored \\
urine.  She has chronic abdominal pain related \\
to her cancer, but does not report \\
any change in her abdominal pain. She has never\\
had an episode like this before.In the ED, \\
initial vitals were T 99.8, HR 95, BP 180/106, \\
RR 18, 99\% on RA.  Her labs were notable \\
for leukocytosis, elevated  LFTs with an \\
obstructive pattern.  Her [**Location (un) **] u/s \\
showed a common bile duct dilation with obstruction \\
at the level of a large pancreatic mass.  She was \\
given Cipro/Flagyl and morphine for pain. ERCP \\
was consulted in the ED and plans to peform \\ 
ERCP on  [**2907-3-3**]. Upon arrival to \\
the [**Hospital Unit Name 2**], she was in \\
no acute distress.
\\
\textbf{Past Medical History:}\\
1.  Metastatic pancreatic cancer with liver mets, \\
diagnosed [**10-11**].  She failed gemcitabine.  \\
She is currently enrolled in hospice but is also \\
interested in considering treatment options.\\
2.  Lupus - no current treatment\\
3.  Hypertension.\\
4.  Hypercholesterolemia.\\
5.  GERD.\\
6.  Hypothyroidism.\\
7.  COPD.\\
8.  History of prior TIAs. \\
…
    }
    &
    \shortstack[l]{
    \textbf{Service:} MEDICINE \\
\textbf{Allergies:} Sulfa, latex, pcn and shellfish \\ \\
\textbf{Major Surgical or Invasive Procedure:} \\
Sphincterotomy \\ \\ \\
\textbf{History of Present Illness:} \\
77 yo female with history of poor \\ 
a[** Location **]e, 2 days of jaundice. \\
No fevers noted. \\\
[**Location \\ (un) **] colored stool. \\
Unchanged chronic abdominal pain. \\
99.8, HR95, BP 180/106. 99\% on RA. \\
WBC of 16.9, tbili 6.2, alk p 461, RUQ U/S \\
with CBD dilation and obstruction at level \\
of pancreatic  mass. Pt. admitted to \\
the micu for further management. \\
\\ \\ \\ \\ \\ \\ \\ \\ \\ \\ \\ \\
\\ \\ \\ \\ \\ \\ \\ \\ \\ \\ \\ \\
\\ \\ \\ \\ \\ \\ \\ \\ \\ \\ \\ \\
\\ \\ \\ \\ \\ \\ \\ \\ \\ \\ \\ \\
\\ \\ \\ \\ \\ \\ \\ \\
\textbf{Past Medical History:}\\ 
 Metastatic Pancreatic Cancer \\
- [**Last Name (NamePattern1) **] \\
- Hyperlipidemia, hypercholesterolemia \\
- afebrile \\
- Leukocytosis \\
- Pancreas cancer \\
- metastatic pancreatitis \\
- malignant neoplasm 
\\
\\
\\
\\
\\
\\
\\
\\
…}
    \\
    \bottomrule
  \end{tabular}
  \caption{Example of T5’s Generated Text using data Setup 1. The actual Discharge Summary Report is on the left side of this table}~\label{tab:eval1}
\end{table*}

\begin{table*}
  \centering
  \begin{tabular}{p{0.5\textwidth} | p{0.4\textwidth}} 
    \toprule
    Actual Discharge Summary Report & Generated Text \\
    \midrule
    \shortstack[l]{
        Mr. [**Known patient lastname 35644**] is a 28 year \\ old man from [**Country **]. He states had a fall \\ from a skateboard at 6 am [**2784-10-5**]  on his \\ way to work (fishing company employee and some \\
        question as to if infection happened at \\
        work and was exercabated by fall).\\ …\\
        Keep your incisions clean and dry\\ Please take all medication as prescribed\\ If you have any increased redness, drainage, \\
        or swelling, or \\…\\Please take only as directed and do not drive or \\
        operate any machinery while taking this medication.\\
        There is a 72 hour (Monday
        through Friday,
   \\ 9am to 4pm) 
    response time for prescription \\
    refil
    requests.
    }
    &
    \shortstack[l]{
    Mr. [**Known patient lastname **] is a 28 \\ yo
    male from [**Country **] who presented \\
    to [**Hospital **] Hospital with c/o fall from \\
    skateboard at 6 am
    Friday on his way to \\
    work (fishing company employee).
    \\ \\ \\ \\ \\ \\ … \\ \\ \\ \\ \\ \\ \\  \\ \\ \\
    Please shower daily including washing \\
    incisions gently with
    mild soap, \\
    no baths or swimming \\
    Monitor wounds \\
    No driving for 1 month \\
    Please call with any questions \\
    or concerns [**Telephone/Fax (1) 181**]
    }
    \\
    \bottomrule
  \end{tabular}
  \caption{Example of T5’s Generated Text using data Setup 5. The actual Discharge Summary Sections are on the left side of this table}~\label{tab:eval2}
\end{table*}

\begin{figure}[]
\begin{center}
  \includegraphics[width=0.95\columnwidth]{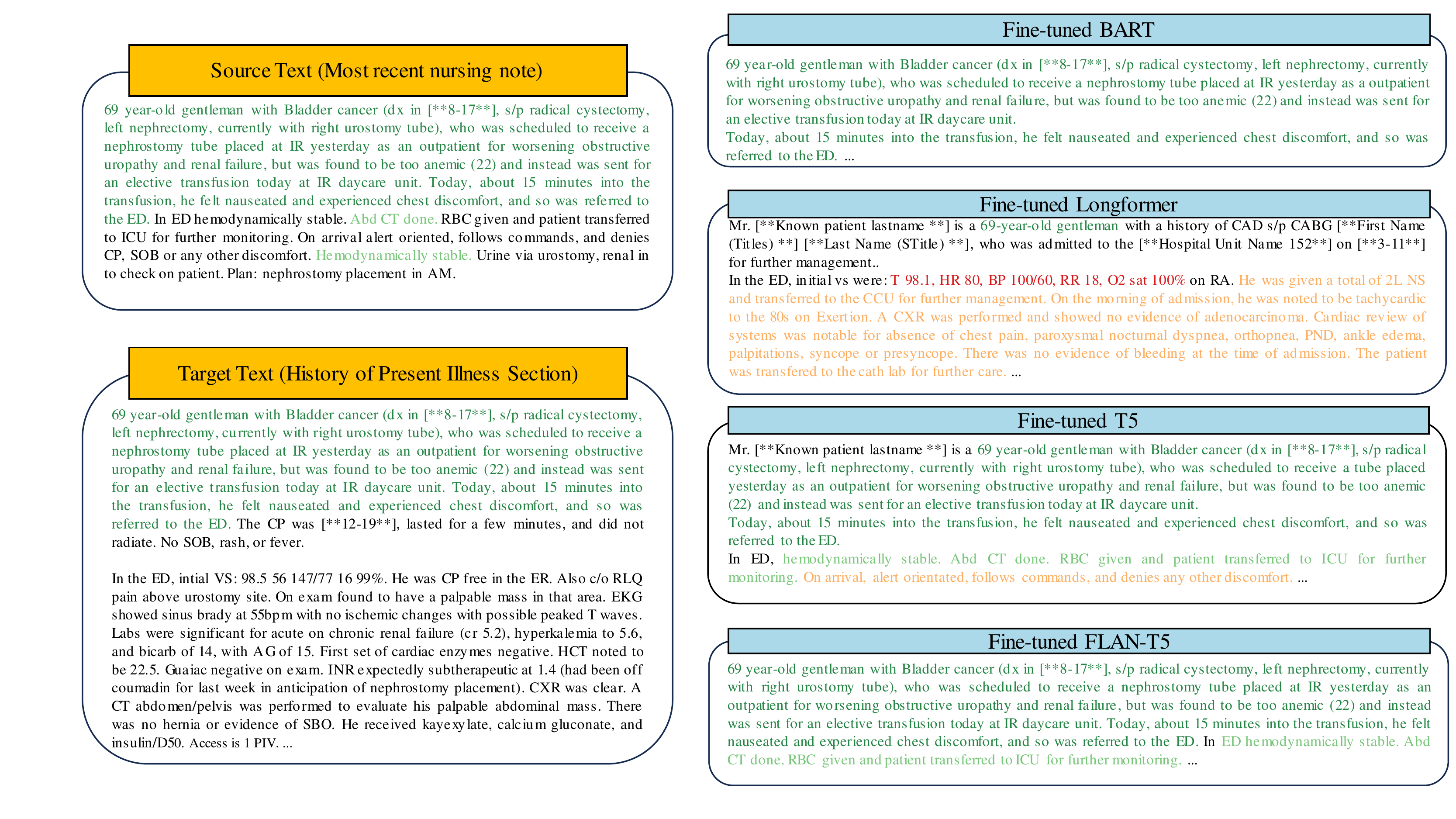}
  \caption{\textbf{History of Present Illness Section Generation}: For each model that was fine-tuned to generate the `History of Present Illness' section, we qualitatively compare one of their test outputs amongst each other. {\color[RGB]{35,132,67} Dark green} matches the target text, i.e., History of Present Illness section. {\color[RGB]{120,198,121} Light green} represents text that match with the source/input text, but not the target text. {\color[RGB]{253,141,60} Orange} text is not present in both source and target text.
  Lastly, {\color[RGB]{227,26,28} Red} text is categorically present in the target text, but does not match in number, detail, or other specifics.
  }~\label{fig:qual_1}
 \end{center}
\end{figure}

\subsection*{Qualitative Analysis}
In Table~\ref{tab:eval1}, we illustrate an instance of a generated summary (on the right side of the table), which can be compared against the actual report (on the left side of the table). This example was taken from one of the generated test outputs by T5 from setup 1. As we can see, this model is able to recognize and identify that the Discharge Summary has a structure with some sections such as Service, Allergies, Past Medical History and more. It was also able to correctly determine the related information for age, service, and allergies. It also recognizes that certain numbers are important. For instance, it kept some lab result display results such as 99.8, HR 95, and BP 180/160. However, there are other instances where the service type or other information can be wrong. To understand how often an information is accurate, we calculate the accuracy of information for a particular section, namely “Service.” Amongst the model’s generated texts, 508 out of 1000 have “Service” sections described. 282 out of 508 (55\%) have the right service type described. This implies that even if the model recognizes this section, it may not always identify the actual type of service conducted on the patient. These behavior traits also appear in other setup and model combinations.

Table~\ref{tab:eval2} takes a closer look at an example taken from one of the generated test outputs by the highest performing model for setup 5, which is T5. This instance illustrates that the model was able to determine the content of two sections, i.e., ``History of Present Illness'' and ``Discharge Instruction'' as intended with this particular data setup. While both sections are generally accurate, the latter section has general instructions that may or may not represent the patient’s actual discharge instructions. Nevertheless, this example shows promise with respect to generating sections within the Discharge Summary Report.

In Figure~\ref{fig:qual_1}, we compare instances of generated summaries by each of the fine-tuned models, which were fine-tuned to generate ``History of Present Illness section." We color sentences that match the source text and/or the target text to visually understand how much overlap there is amongst each text. Between the input (source) and the reference output (target), the former is almost entirely present in the latter's content. This implies that nursing notes do have content that is ultimately utilized in the discharge summary report. As for the model outputs, BART, T5, and FLAN-T5 recognize such content overlap behavior as it copies the most of the source output as part of its generated summary. Amongst all model outputs, Longformer seems to hallucinate the most since we find information that is present in neither of the source and target texts. 

\section*{DISCUSSION}
\subsection*{Clinical Importance and Implications}
Hospital discharge summaries are a key source of patient information. Writing good discharge summaries requires considerable provider resources. Without automated generation, healthcare providers will continue to face a growing burden of documentation resulting in delayed report generation as well as decreased face-to-face interaction with patients. Our study benchmarks various language model and their data setups to find those settings that automatically generate key sections of this document with decent ROUGE scores. Based on the results presented in the previous section, we conclude that the FLAN-T5 can generate the discharge summary with the highest ROUGE score compared across all models and setups. As setups 1, 2, and 3 share the same generation target, their corresponding ROUGE performance metrics can be directly compared. This implies that setup 1 is on average a better direction to follow as compared with setups 2 and 3. Moreover, we observe that, contrary to the common perception in standard summarization tasks such as news summarization were lead-3 is a very strong baseline, for EHR summarization as tested in setup 3, the lead-3 baseline does not perform well when training a model on it and testing that model against the ground-truth discharge summaries. That is a major difference between summarization of EHRs compared with other types of documents such as news articles where the first few sentences of a document present the main gist of the document. Furthermore, we observed in our experiments in setups 4 and 5 the viability of generation of history of present illness and the discharge instructions with a high quality in terms of ROUGE performance.

\clearpage
\subsection*{Applying models in practice}
As shown in the results section, the generated text output was not always matching to the actual circumstance. Since these models have been experimented on with the intent to summarize medical documents, their ROUGE scores indicate promising discharge summaries. However, there is a need for further assessment in terms of factual correction in all instances. They are generally accurate, but there are still special cases in which the model's inference is not fully correct. Hence, we should look for ways to add other metrics or algorithms to produce factually-correct information more frequently. By doing so, we can learn to trust the generation process more than we do now.

\subsection*{Limitations and Future Extensions}

In the experiment setup, we solely focused on nursing notes as inputs to models for producing discharge summaries. However, the MIMIC-III database contains other types of provider documentation - such as physician notes - that may be used in the construction of discharge summaries. Hence, as future extensions, we can explore if a combination of these notes or some other sets of text improves the overall discharge summary generation with more factual correction. Furthermore, we plan to extend this work to generate topic-based~\cite{cats, newts} summaries focusing on each organ at a time to construct a coherent summary.

\section*{CONCLUSIONS}

This article benchmarks various training regimes and models to automatically generate EHR Discharge Summaries.
In terms of data training setup, we find that utilizing full nursing notes (setup 1) and focusing on generating specific sections (setups 4 and 5) can allow consistent improvement in text summarization by most language models, especially for models such as BART and T5. Amongst the pre-trained models, we find that FLAN-T5 can be more reliable to produce unseen EHR summarizations. In future work, we encourage the research community to continue its collaboration with medical professionals to create summarization-based medical datasets and even better summarization models to further enable more readable and accurate medical documents.


\clearpage
\clearpage
\section*{FUNDING STATEMENT}
This research is supported in part by the SNSF (P2TIP2\_187932) and grant T32DA013911 from the National Institute on Drug Abuse, of the NIH. The views and conclusions contained herein are those of the authors and should not be interpreted as necessarily representing the official policies, either expressed or implied, of SNSF, or NIH.

\section*{COMPETING INTERESTS STATEMENT}
The authors have no competing interests to declare.

\section*{CONTRIBUTORSHIP STATEMENT}
C.E. and S.B. contributed to the research idea. K.P. and S.B. worked on designing the methodology and experiments. K.P. implemented the data processing, modeling, and data analysis. K.P., S.B., L.M., and C.E. discussed the results and contributed to the final manuscript.

\section*{DATA AVAILABILITY}
The data for this research was gathered from the MIMIC-III dataset. The data organized for all the setups in this article will be released.

\printbibliography
\clearpage

\clearpage
\newpage
\renewcommand{\thesection}{S\arabic{section}}
\renewcommand{\thefigure}{S\arabic{figure}}
\renewcommand{\thetable}{S\arabic{table}}
\setcounter{section}{0}
\setcounter{figure}{0}
\setcounter{table}{0}

\section*{Summarization and Generation of Discharge Summary Medical Reports}

\section*{Supplementary Material}
\label{section:supp-materials}
\section{Technical Implementation}

All the training and testing have been conducted using Python notebooks. The models use Hugging Face packages\footnote{\href{https://huggingface.co/}{Hugging Face Website}}. For BART, we use BartForConditionalGeneration and BartTokenizer classes\footnote{\href{https://huggingface.co/docs/transformers/model_doc/bart}{Bart-related Hugging Face documentation}}. We utilize transfer learning by starting our model training using the values present in the “facebook/bart-base” model\footnote{\href{https://huggingface.co/facebook/bart-base}{Bart-base model card link}}. For T5, we useT5ForConditionalGeneration and T5ForConditionalGeneration classes\footnote{\href{https://huggingface.co/docs/transformers/model_doc/t5}{T5-related  Hugging Face documentation}}. Similar to BART, we fine-tune a pre-trained model, which for T5 is the “t5-base” model\footnote{\href{https://huggingface.co/t5-base}{T5-base model card link}}. Similarly, we fine-tune FLAN-T5 from the pre-trained model, ``google/flan-t5-base"\footnote{\href{https://huggingface.co/google/flan-t5-base}{FLAN-T5-base model card link}}. For Longformer, we use the variant known as Longformer-Encoder-Decoder (LED), which is mentioned in the Longformer paper~\cite{longformer}. This variant is suitable for summarization tasks as it has the support for performing long document generative sequence-to-sequence tasks. Hence the Hugging Face classes that we use for LED are LEDForConditionalGeneration and LEDTokenizer\footnote{\href{https://huggingface.co/docs/transformers/main/en/model_doc/led}{LED-related Hugging Face documentation}}. As for the pre-trained model, we use “allenai/led-base-16384”\footnote{\href{https://huggingface.co/allenai/led-base-16384}{LED-base-16384 model card link}}. We chose our pre-trained models based on the high number of downloads recorded by Hugging Face.

In each of these models, there are a couple of parameters we needed to change in order to maximize our summary generation accuracy. One of such parameters is known as \texttt{num\_beams}. This parameter represents the number of beams (or possibilities) that the model will use while conducting a beam search, which is a method to reduce the likelihood of missing a highly probable word sequence: the model will keep \texttt{num\_beams} number of hypotheses at each time step and eventually choose the hypothesis that gives the overall highest word sequence probability. Hence, to determine this number, we trained and tested one of the models, BART, with one of our setups, Setup 3, with beam values 1 through 10. The set of metrics we used to test the resulting trained model is known as Recall-Oriented Understudy for Gisting Evaluation (ROUGE)~\cite{lin-2004-rouge}. It works by comparing the generated summary text against the reference summary provided. In Table~\ref{tab:beamsearch}, we display our experimental results where we show how the change in \texttt{num\_beams} value affects the ROUGE scores for the BART model, which is trained using Setup 3. Based on the results, the \texttt{num\_beams} value 10 provides the highest value in the entire set of ROUGE scores. Hence, when we generate our summaries, the parameter value we assign our \texttt{num\_beams} to is 10.

\begin{table}
  \centering
  \begin{tabular}{l r r r r}
    \toprule
    Number of Beams & & \multicolumn{2}{c}{\small{Chosen Setup No. 3}} \\
    \cmidrule(r){2-5}
    &
    {\small\textit{rouge1}}
    & {\small \textit{rouge2}}
      & {\small \textit{rougeL}}
    & {\small \textit{rougeLsum}} \\
    \midrule
    1 & 21.4 & 17.5 & 20.6 & 20.6 \\
    2 & 21.8 & 17.8 & 21.0 & 21.0 \\
    3 & 22.0 & 18.0 & 21.2 & 21.2 \\
    4 & 21.9 & 17.9 & 21.0 & 21.1 \\
    5 & 22.0 & 18.0 & 21.2 & 21.2 \\
    6 & 22.1 & 18.0 & 21.2 & 21.3 \\
    7 & 22.1 & 18.0 & 21.2 & 21.2 \\
    8 & 22.0 & 18.0 & 21.3 & 21.2 \\
    9 & 22.1 & 18.1 & 21.3 & 21.2 \\
    \textbf{10} & \textbf{22.1} & \textbf{18.1} & \textbf{21.3} & \textbf{21.3} \\
    \bottomrule
  \end{tabular}
  \caption{Beam Search Range}~\label{tab:beamsearch}
\end{table}

In our scripts, we calculate ROUGE using the package provided by Hugging Face which is a wrapper around Google Research’s re-implementation of ROUGE\footnote{\href{https://huggingface.co/metrics/rouge}{Hugging Face ROUGE}}. 

\end{document}